%% file: 01-main.tex
\pdfoutput=1
\documentclass[conference]{IEEEtran}
\IEEEoverridecommandlockouts

\usepackage{cite}
\usepackage{amsmath,amssymb}
\usepackage{graphicx}
\usepackage{booktabs}
\usepackage{multirow}
\usepackage{enumerate}
\usepackage{tabularx}
\usepackage{url}
\usepackage[caption=false,font=footnotesize]{subfig}

\usepackage{stfloats} 

\begin{document}

\title{MoCLIP-Lite: Efficient Video Recognition by Fusing CLIP with Motion Vectors}

\author{Binhua Huang\textsuperscript{1,2}, 
    Ni Wang\textsuperscript{3},
    Arjun Pakrashi\textsuperscript{2,4},
    Soumyabrata Dev\textsuperscript{1,2,4}
	\vspace{0.5em}\\
	\textsuperscript{1}The ADAPT SFI Research Centre, Dublin, Ireland\\
    \textsuperscript{2}School of Computer Science, University College Dublin, Ireland\\
    \textsuperscript{3}Amazon Development Center Germany GmbH, Berlin, Germany\\
    \textsuperscript{4}Beijing-Dublin International College, Beijing, China\\
    bh.huang@ieee.org, niwang.fr@gmail.com, arjun.pakrashi@ucd.ie, soumyabrata.dev@ucd.ie
    \thanks{This research was conducted with the financial support of Research Ireland under Grant Agreement No.\ 13/RC/2106\_P2 at the ADAPT Centre at University College Dublin. ADAPT, the Research Ireland Centre for AI-Driven Digital Content Technology, is funded through the Research Ireland Centres Programme.}
}

\maketitle

\input{sec/0_abs}

\input{sec/1_intro}

\input{sec/3_method}

\input{sec/4_exp}

\input{sec/5_con}


\bibliographystyle{IEEEtran}
\bibliography{refs}
\end{document}

%% file: sec/0_abs.tex
\begin{abstract}
Video action recognition is a fundamental task in computer vision, but state-of-the-art models are often computationally expensive and rely on extensive video pre-training. In parallel, large-scale vision-language models like Contrastive Language-Image Pre-training (CLIP) offer powerful zero-shot capabilities on static images, while motion vectors (MV) provide highly efficient temporal information directly from compressed video streams. To synergize the strengths of these paradigms, we propose MoCLIP-Lite, a simple yet powerful two-stream late fusion framework for efficient video recognition. Our approach combines features from a frozen CLIP image encoder with features from a lightweight, supervised network trained on raw MV. During fusion, both backbones are frozen, and only a tiny Multi-Layer Perceptron (MLP) head is trained, ensuring extreme efficiency. Through comprehensive experiments on the UCF101 dataset, our method achieves a remarkable 89.2\% Top-1 accuracy, significantly outperforming strong zero-shot (65.0\%) and MV-only (66.5\%) baselines. Our work provides a new, highly efficient baseline for video understanding that effectively bridges the gap between large static models and dynamic, low-cost motion cues. Our code and models are available at \url{https://github.com/microa/MoCLIP-Lite}.
\end{abstract}


%% file: sec/1_intro.tex
\section{Introduction}
\label{sec:introduction}

Video action recognition stands as a fundamental challenge in computer vision, with a wide array of applications ranging from intelligent surveillance and human-computer interaction to robotics and autonomous systems. The primary goal is to enable machines to understand and interpret the dynamic content within video streams. In recent years, significant progress has been driven by deep learning models, particularly deep Convolutional Neural Networks (CNN) and Transformers \cite{carreira2017quo, bertasius2021space, arnab2021vivit}. These heavyweight models have achieved state-of-the-art (SOTA) performance on benchmark datasets. However, their success often comes at the cost of immense computational complexity and a heavy reliance on large-scale video datasets like Kinetics \cite{kay2017kinetics} for pre-training, creating a significant barrier for applications in resource-constrained environments.

To mitigate these challenges, research has also explored more efficient avenues for video understanding. A classic and effective approach is the two-stream network \cite{simonyan2014two}, which processes appearance (RGB frames) and motion (optical flow) in parallel pathways, demonstrating the value of treating these modalities separately. Building on this principle of efficiency, another line of work has focused on leveraging information directly from the compressed video domain, bypassing the costly decoding process entirely. Methods like CoViAR \cite{wu2018compressed} have shown that motion vectors (MV), readily available in compressed streams like H.264/AVC, can serve as a powerful and highly efficient source of motion information. While these methods are computationally inexpensive, they often lack the rich semantic understanding of object and scene context that large-scale pre-trained models provide. More recently, large-scale vision-language models (VLM) like Contrastive Language-Image Pre-training (CLIP) \cite{radford2021learning} have revolutionized image understanding with their remarkable zero-shot generalization capabilities. However, effectively and efficiently adapting these powerful static image models to the temporal dynamics of video remains an open and active area of research.

To study this efficiency–semantics trade-off in a standardized setting, we adopt the Temporal Segment Network (TSN) formulation \cite{wang2018temporal}, which sparsely samples segments across a video to capture long-range dynamics with modest cost and naturally accommodates compressed-domain inputs such as motion vectors. We evaluate computational efficiency using floating-point operations (FLOPs), reporting \emph{total} inference compute under each method’s own test-time view protocol (i.e., per-view FLOPs multiplied by the number of temporal segments and spatial crops) to enable fair comparison. Unless otherwise specified, experiments are conducted on the widely used UCF101 benchmark \cite{soomro2012ucf101}, which provides a common ground for both compressed-domain and vision–language baselines.

In this paper, we bridge the gap between these paradigms. We demonstrate that the semantic richness of CLIP and the computational efficiency of motion vectors are highly complementary. 
We propose MoCLIP-Lite, a simple yet powerful late-fusion framework that integrates a frozen, pre-trained CLIP image encoder with a lightweight supervised motion-vector network. 
Our work makes the following contributions:
\begin{itemize}
    \item We present a novel two-stream late fusion architecture that effectively integrates pre-computed semantic features from CLIP with dynamically extracted motion features from a TSN-based network operating directly on motion vectors.
    \item We demonstrate the extreme efficiency of our approach. During the fusion stage, we freeze both the CLIP encoder and the pre-trained motion vectors encoder, and only train a tiny MLP head (0.97M parameters), our method achieves a massive performance gain with a negligible increase in inference cost (only 0.05 GFLOPs over the CLIP-only + MV-only baseline).
    \item Through experiments on the UCF101 dataset, we show that MoCLIP-Lite achieves a remarkable 89.2\% Top-1 accuracy, significantly outperforming both the strong zero-shot CLIP baseline (65.0\%) and the fully supervised MV-only baseline (66.5\%). Our in-depth analysis further reveals the strong complementarity between the appearance and motion modalities.
\end{itemize}

%% file: sec/3_method.tex
\section{Methodology}
\label{sec:methodology}

\begin{figure*}[ht!]
    \centering
    \includegraphics[width=1\linewidth]{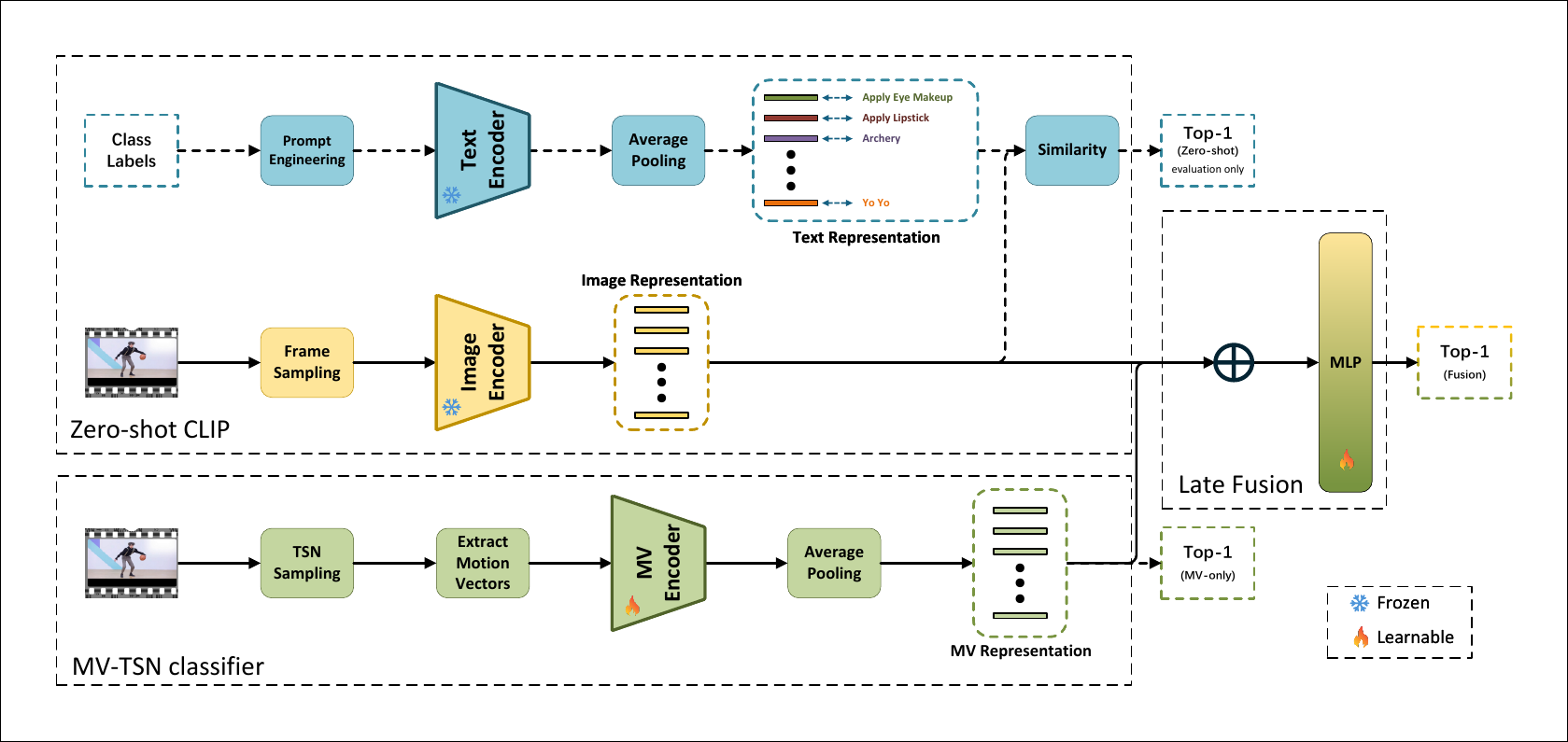} 
    \caption{Overview of MoCLIP-Lite and its three evaluated components. \textbf{(A) Zero-shot CLIP:} Offline, we build a class text library via prompt engineering with the frozen CLIP text encoder, and obtain a pre-computed appearance embedding for each video using the frozen CLIP image encoder (also giving a zero-shot baseline). \textbf{(B) MV-TSN classifier:} Motion features are extracted from compressed-domain motion vectors with a lightweight TSN-based MV-encoder; this branch can be evaluated as an MV-only baseline. \textbf{(C) Late fusion:} Online, the pre-computed appearance embedding is concatenated with the motion vectors features and passed to a lightweight MLP head to predict the final label; both encoders are frozen, and only the MLP is trainable. (Snowflake/flame icons denote frozen/learnable parameters.)}
    \label{fig:architecture} 
\end{figure*}

To address the task of efficient video action recognition, we propose a two-stream late fusion framework. Our approach is designed to synergistically combine the powerful semantic understanding of a large-scale, pre-trained vision-language model with the rich dynamic information captured by motion vectors. The overall architecture, illustrated in Figure~\ref{fig:architecture}, consists of an appearance pathway and a motion pathway, whose features are concatenated and processed by a lightweight classification head.  

\noindent\textbf{Figure~\ref{fig:architecture} (Overview).}
Figure~\ref{fig:architecture} summarizes MoCLIP-Lite and the three evaluation settings we study.
\emph{(A) Zero-shot CLIP (offline).} We first construct a class text library via prompt ensembling with the \emph{frozen} CLIP text encoder, and obtain for each video a pre-computed appearance embedding using the \emph{frozen} CLIP image encoder; this branch also provides a strong zero-shot baseline by matching image and text features.
\emph{(B) MV-TSN classifier (supervised).} Motion cues are extracted directly from compressed-domain motion vectors with a lightweight TSN-style MV-encoder; this branch can be evaluated independently as our MV-only baseline.
\emph{(C) Late fusion (online).} At inference/training for fusion, the pre-computed appearance embedding from (A) is concatenated with the motion vectors feature from (B) and fed to a small MLP head to predict the final label. During this stage, \emph{both encoders remain frozen} and only the MLP is trainable, yielding large accuracy gains with negligible extra compute.
The snowflake/flame icons in the diagram denote \emph{frozen} versus \emph{learnable} parameters, respectively.
This overview also reflects our reporting policy: CLIP features are computed once offline, and the online cost is dominated by the motion vectors pathway, while the MLP adds only a tiny overhead.

\subsection{Appearance Pathway: CLIP Feature Extraction}
\label{subsec:appearance}

The appearance pathway serves as a static content encoder, leveraging the robust generalization capabilities of the pre-trained Contrastive Language-Image Pre-training (CLIP) model \cite{radford2021learning}. We use the official CLIP model with a Vision Transformer (ViT-B/32) \cite{dosovitskiy2020image} backbone as our frozen image encoder, denoted as $\mathcal{E}_{I}$.

For a given video $V = \{I_1, I_2, \dots, I_T\}$, we sample a single representative RGB frame $I_{rep}$ to capture its global scene content. To establish a strong and fair zero-shot baseline, we employ prompt engineering. A set of $K$ descriptive templates (e.g., "a photo of a person {}") are used to generate multiple text prompts for each action class. The features for these prompts are then averaged to create a single, robust text embedding for each class.

To maximize training efficiency for MoCLIP-Lite, we pre-compute the appearance features for all videos. The 512-dimensional appearance feature vector $\mathbf{f}_{\text{app}} \in \mathbb{R}^{d_{\text{clip}}}$ is generated as:
$$
\mathbf{f}_{\text{app}} = \mathcal{E}_{I}(\text{Preprocess}(I_{rep}))
$$
where $d_{\text{clip}}=512$. These pre-computed features are stored on disk and loaded directly during the fusion training stage, avoiding any repeated computations from the CLIP encoder.  

While using a single frame does not capture the full temporal evolution, it provides a highly efficient yet powerful representation of the video's global scene and object context, serving as a strong semantic prior for MoCLIP-Lite.  

\emph{Prompt Templates.}
Following CLIP~\cite{radford2021learning}, we adopt prompt ensembling for each class and
release the exact templates and class-name mappings in our repository; per-class text embeddings are obtained by averaging all prompts.

\subsection{Motion Pathway: TSN-based motion vectors Feature Extraction}
\label{subsec:motion}

The motion pathway is designed to learn discriminative representations from the temporal dynamics of an action.

\paragraph{Input Representation} We utilize raw motion vectors extracted directly from the compressed video stream using the CoViAR library. This avoids the costly step of video decoding. The motion vectors are represented as a 2-channel tensor $M_t \in \mathbb{R}^{H \times W \times 2}$ ($MV_x, MV_y$) and are normalized by a function $\mathcal{N}_{\text{mv}}$ into a range of $[-0.5, 0.5]$.  
The normalization function $\mathcal{N}_{\text{mv}}$ is critical for mapping the raw motion vectors into a distribution suitable for the pre-trained backbone. Specifically, the raw pixel displacements are first scaled by a factor of $127.5/20$ to map a typical motion range of $[-20, 20]$ pixels to $[-127.5, 127.5]$. An offset of 128 is then added to shift the range to approximately $[0, 255]$. Finally, this 8-bit representation is linearly scaled to $[-0.5, 0.5]$ before being fed into the network. This process is consistent with prior work on MV-based action recognition \cite{wu2018compressed}.  

\paragraph{Temporal Sampling} We adopt the Temporal Segment Networks (TSN) \cite{wang2018temporal} sampling strategy. The video's motion vectors stream is divided into $N$ uniform segments, and one motion vectors frame $M_{t_k}$ is sampled from each segment $k$. In our experiments, we use $N=3$.  
It is important to note the distinction between our training and testing sampling strategies. During training, we use \( N_{\text{train}} = 3 \) segments and randomly pick one motion vectors frame per segment. This random sampling serves as a form of data augmentation, preventing the model from overfitting to specific temporal locations. During testing, we follow a multi-view protocol with \( N_{\text{test}} = 32 \) temporal segments and select the center frame in each segment (single center crop per view). Predictions are averaged over the 32 views. This matches the view protocol used in our accuracy and FLOPs reporting.

\paragraph{Backbone Architecture} We employ an EfficientNet-B0 \cite{tan2019efficientnet}, pre-trained on ImageNet, as the backbone for the motion feature extractor, denoted as $\mathcal{E}_{M}$. The model is adapted for 2-channel input by replacing its first convolutional layer and initializing its weights from the original 3-channel weights. An additional "nn.BatchNorm2d(2)" layer is prepended to normalize the motion vectors data distribution. The final motion feature for the video, $\mathbf{f}_{\text{motion}} \in \mathbb{R}^{d_{\text{mv}}}$, is obtained by averaging the features from all segments:
$$
\mathbf{f}_{\text{motion}} = \frac{1}{N} \sum_{k=1}^{N} \mathcal{E}_{M}(\mathcal{N}_{\text{mv}}(M_{t_k}))
$$
where $d_{\text{mv}}=1280$. This motion pathway is first trained independently to establish the MV-only baseline.  

We selected EfficientNet-B0 as the backbone due to its excellent trade-off between performance and computational efficiency, aligning with the overall goal of our lightweight framework.  

\subsection{Multimodal Late Fusion Framework}
\label{subsec:fusion}

We opt for a late fusion strategy because it allows each specialized backbone (CLIP and MV-encoder) to process its native modality independently, preserving their powerful representations. This modular approach is simpler to implement and more effective than early fusion, which would require reconciling disparate input shapes (e.g., 3-channel RGB and 2-channel motion vectors).  

The final stage of our framework fuses the unimodal features from the two pathways into a powerful multimodal representation.   

\paragraph{Feature Concatenation.} The final representation for a video is formed by concatenating the pre-computed appearance feature $\mathbf{f}_{\text{app}}$ and the aggregated motion feature $\mathbf{f}_{\text{motion}}$:
$$
\mathbf{f}_{\text{fusion}} = [\mathbf{f}_{\text{app}} \, ; \, \mathbf{f}_{\text{motion}}]
$$
where $[\cdot \, ; \, \cdot]$ denotes concatenation, resulting in a fused vector $\mathbf{f}_{\text{fusion}} \in \mathbb{R}^{d_{\text{clip}} + d_{\text{mv}}}$.

\paragraph{MLP Classification Head \& Training Strategy.} This fused vector is passed to a lightweight Multi-Layer Perceptron (MLP) head, $\mathcal{C}_{\text{fusion}}$, for final classification. The final prediction is given by $\mathbf{y} = \mathcal{C}_{\text{fusion}}(\mathbf{f}_{\text{fusion}})$. Our training strategy is highly efficient: during this fusion stage, both the CLIP image encoder and the pre-trained motion feature extractor $\mathcal{E}_{M}$ (with weights loaded from the best MV-only model) are frozen. Only the parameters of the MLP head $\mathcal{C}_{\text{fusion}}$ are trained using a standard cross-entropy loss, $\mathcal{L}_{\text{CE}}$. This allows the model to learn the optimal combination of the two expert features with minimal computational cost.  
The MLP head is intentionally designed to be lightweight. The input layer reduces the 1792-dimensional concatenated feature vector ($1280$ from EfficientNet-B0 and $512$ from CLIP) to a 512-dimensional hidden representation. This is followed by a ReLU activation and a Dropout layer with a rate of 0.5 for regularization, before the final linear layer produces logits for the 101 classes.

%% file: sec/4_exp.tex
\section{Experiments and Results}
\label{sec:experiments}

In this section, we present a comprehensive evaluation of our proposed late fusion framework. 
We first detail the experimental setup, including the dataset, evaluation metrics, and implementation details. 
summary of the overall comparison with state-of-the-art methods is provided in Table~\ref{tab:sota_comparison}. 
We then compare our final MoCLIP-Lite model against a range of state-of-the-art methods. 
Finally, we provide an ablation study and in-depth analyses to demonstrate the effectiveness of our components and explore capabilities of the model.  

\begin{table*}[!t]
\centering
\caption{Comprehensive comparison on UCF101. Top-1 accuracies are reported. FLOPs denote total inference compute under testing protocol of each method.}
\label{tab:sota_comparison}
\resizebox{\textwidth}{!}{%
\begin{tabular}{l l l c c c}
\toprule
\textbf{Method} & \textbf{Modality} & \textbf{Architecture} & \textbf{Trainable Params (M)} $\downarrow$ & \textbf{Inference GFLOPs} $\downarrow$ & \textbf{Top-1 Acc. (\%)} $\uparrow$ \\
\midrule
\multicolumn{6}{@{}l}{\textit{\textbf{Zero-Shot VLM}}} \\
\midrule
CLIP (ViT-B/32) \cite{radford2021learning} & RGB (zero-shot) & ViT-B/32 & 0 & 4.4 & 64.5 \\
X-CLIP \cite{ni2022expanding} & RGB (zero-shot) & ViT-B/16 & N/G & 1732 & 72.0$^{\S}$ \\
EPK-CLIP \cite{yang2024epk} & RGB (zero-shot) & ViT-B/16 & N/G & 1732 & \textbf{75.3}$^{\S}$ \\
\textbf{CLIP-only (Ours)} & \textbf{RGB (zero-shot)} & \textbf{ViT-B/32} & \textbf{0} & \textbf{4.4} & 65.0 \\
\midrule
\multicolumn{6}{@{}l}{\textit{\textbf{MV-only methods}}} \\
\midrule
CoViAR (MV branch) \cite{wu2018compressed} & MV-only & ResNet-18 & 11.7 & 44.5$^{\dagger}$ & 63.9 \\
EMV-CNN \cite{zhang2016real} & MV-only & ClarifaiNet & 61$^{\dagger}$ & 65$^{\dagger}$ & 79.3 \\
DTMV-CNN \cite{zhang2018real} & MV-only & ClarifaiNet & 61$^{\dagger}$ & 65$^{\dagger}$ & \textbf{80.3} \\
\textbf{MV-only (Ours)} & \textbf{MV-only} & \textbf{EfficientNet-B0} & \textbf{5.3} & \textbf{12.5} & 66.5 \\
\midrule
\multicolumn{6}{@{}l}{\textit{\textbf{Fusion methods}}} \\
\midrule
Two-Stream \cite{simonyan2014two} & RGB + Flow & VGG-M + VGG-M & 91.2$^{\dagger}$ & 83$^{\dagger}$ & 86.2 \\
DTMV+RGB-CNN \cite{zhang2018real} & RGB + MV & ClarifaiNet$\times$2 & 122 & 83$^{\dagger}$ & 87.5$^{\S}$ \\
\textbf{MoCLIP-Lite (Ours)} & \textbf{RGB + MV} & \textbf{ViT-B/32 + EffNet-B0} & \textbf{0.97} & \textbf{16.9} & \textbf{89.2} \\
\bottomrule
\end{tabular}%
}
\vspace{4pt}
\begin{minipage}{\textwidth}\footnotesize
\textbf{Notes.} 
$\uparrow$ higher is better; $\downarrow$ lower is better.
$^{\dagger}$ Not reported by the original paper; estimated by us.
$^{\S}$ All other accuracies are reported on UCF101 Split~1. This mark indicates results from papers that only report 3-split averages. 
\noindent\emph{Reporting policy.}
For fairness, we count total FLOPs as (per-view FLOPs)~$\times$~(\#views);
for our fusion this equals motion vectors(32 temporal views, single center crop) $+$ CLIP (1 view) $+$ a negligible MLP head.
\end{minipage}
\end{table*}

\subsection{Experimental Setup}
\label{subsec:expset}

\paragraph{Dataset.}
All experiments are conducted on the widely-used UCF101 dataset \cite{soomro2012ucf101}, which contains 13,320 video clips from 101 action categories. We follow the official protocol and use the provided train/test split 1, which consists of 9,537 training videos and 3,783 testing videos.

\paragraph{Evaluation Metric.}
For the accuracy metric, we report Top-1 classification accuracy (\%) on the test split. For view protocol \& FLOPs accounting, we report total inference FLOPs under each method’s own testing protocol, i.e., per-view FLOPs multiplied by the number of views (temporal clips $\times$ spatial crops). In our models, CLIP-only uses 1 view (a single frame with a center crop), MV-only uses 32 temporal views with a single center crop per view, and MoCLIP-Lite (fusion) sums the costs of both branches plus a small MLP head ($\approx$\,0.05 GFLOPs). Formally,
\resizebox{0.95\linewidth}{!}{%
\ensuremath{
\text{FLOPs}_{\text{total}}
=\sum_{b\in\{\text{CLIP},\,\text{MV}\}}
\text{FLOPs}_{\text{per-view}}^{(b)} \times \#\text{views}^{(b)} \;+\; \text{FLOPs}_{\text{MLP}}
}}

For compared methods, we follow their \emph{published view settings}; when not explicitly stated, we adopt the de-facto settings used in their code or papers. This policy aligns the \emph{reported accuracy} with the \emph{actual compute} required to obtain it.

\paragraph{Implementation Details.}
Our framework consists of two main branches. The appearance pathway uses the official pre-trained CLIP model with a ViT-B/32 image encoder, which is kept frozen throughout all experiments. 

The motion pathway employs an EfficientNet-B0 backbone, pre-trained on ImageNet. For the MV-only baseline, this model is trained for 200 epochs using an Adam optimizer with an initial learning rate of $10^{-2}$ and a weight decay of $10^{-4}$. The learning rate is decayed by a factor of 10 at epochs 80 and 160. During training, we sample 3 segments per video and apply strong data augmentation, including multi-scale cropping and a specialized horizontal flip for motion vectors, as detailed in Section~\ref{subsec:motion}. 

\paragraph{Hardware \& Runtime Setup.}
All experiments are run on a single NVIDIA RTX~3090 (24\,GB) with PyTorch and CUDA.
Unless noted, throughput in Table~II is measured with batch size~1 (per video) in FP32.
We report total inference FLOPs at $224{\times}224$ resolution under each method's own test-time view protocol,
matching the associated accuracy (Sec.~\ref{subsec:expset}).

During late fusion, we initialize the motion backbone with the best-performing MV-only model and keep it frozen across training and inference to preserve its learned representations. 
Only the newly added MLP classification head is trained for 50 epochs using an AdamW optimizer with a learning rate of $10^{-4}$. The final reported accuracies for all supervised models are obtained using a 32-view testing protocol with a single center crop per view. 
For the efficiency comparison in Table \ref{tab:sota_comparison}, all GFLOPs are reported as total FLOPs under the respective testing protocol to ensure a fair comparison of architectural complexity.

\subsection{Comparison with State-of-the-Art Methods}

To provide a fair comparison, Table~\ref{tab:sota_comparison} organizes representative methods into three categories: zero-shot vision–language models (VLMs), motion-vector (MV)-only networks, and fusion approaches.
In the zero-shot setting, large-scale VLMs such as X-CLIP and EPK-CLIP achieve the strongest performance but incur extremely high inference costs (over 1,700 GFLOPs), whereas our CLIP-only baseline reaches a comparable 65.0\% accuracy with just 4.4 GFLOPs.
For MV-only methods, early CNN-based designs (e.g., EMV-CNN, DTMV-CNN) report higher accuracies but rely on heavy backbones with tens of millions of parameters, while our lightweight EfficientNet-B0 model attains 66.5\% accuracy with only 5.3M parameters and 12.5 GFLOPs.
Finally, fusion-based approaches demonstrate the complementary benefits of combining modalities: our MoCLIP-Lite achieves 89.2\% Top-1 accuracy—surpassing both zero-shot VLMs and MV-only methods—while requiring fewer than 1M trainable parameters and just 16.9 GFLOPs, establishing a new state-of-the-art trade-off between efficiency and accuracy for compressed-domain video recognition.

\subsection{Ablation Study and Analysis}

To validate the effectiveness of our fusion strategy and its components, we conduct an ablation study, with results shown in Table \ref{tab:ablation_study}. The CLIP-only and MV-only baselines achieve strong standalone results of 65.0\% and 66.5\% respectively. By simply fusing these two streams with a lightweight MLP, our final model achieves a massive absolute improvement of +22.1\% over the stronger MV-only baseline, clearly demonstrating the strong complementarity between appearance and motion modalities.

\begin{table}[!t]
\centering
\caption{Ablation study of our framework components on UCF101. Throughput is measured on a single NVIDIA RTX 3090 GPU.}
\label{tab:ablation_study}
\resizebox{\linewidth}{!}{%
\begin{tabular}{l c c c}
\toprule
\textbf{Component} & \textbf{Trainable Params (M)} & \textbf{Throughput (videos/sec)} & \textbf{Top-1 Acc. (\%)} \\
\midrule
CLIP-only & 0 & N/A (Offline) & 65.0 \\
MV-only & 5.3 & 286.85 & 66.5 \\
\textbf{Fusion} & \textbf{0.97} & \textbf{275.31} & \textbf{89.2} \\
\bottomrule
\end{tabular}
}
\end{table}

\paragraph{Efficiency Analysis.}
As detailed in Table \ref{tab:ablation_study}, our fusion framework is extremely efficient. The +22.1\% accuracy gain is achieved by training only an additional 0.97M parameters in the MLP head. Furthermore, because the CLIP features are pre-computed, the online inference cost is dominated by the motion pathway. This is reflected in the real-world throughput, which remains high at 275 videos/sec, almost identical to the MV-only model. This demonstrates a substantial performance boost at a minimal cost in both training and inference.

\begin{figure}[!t]
    \centering
    \includegraphics[width=\linewidth]{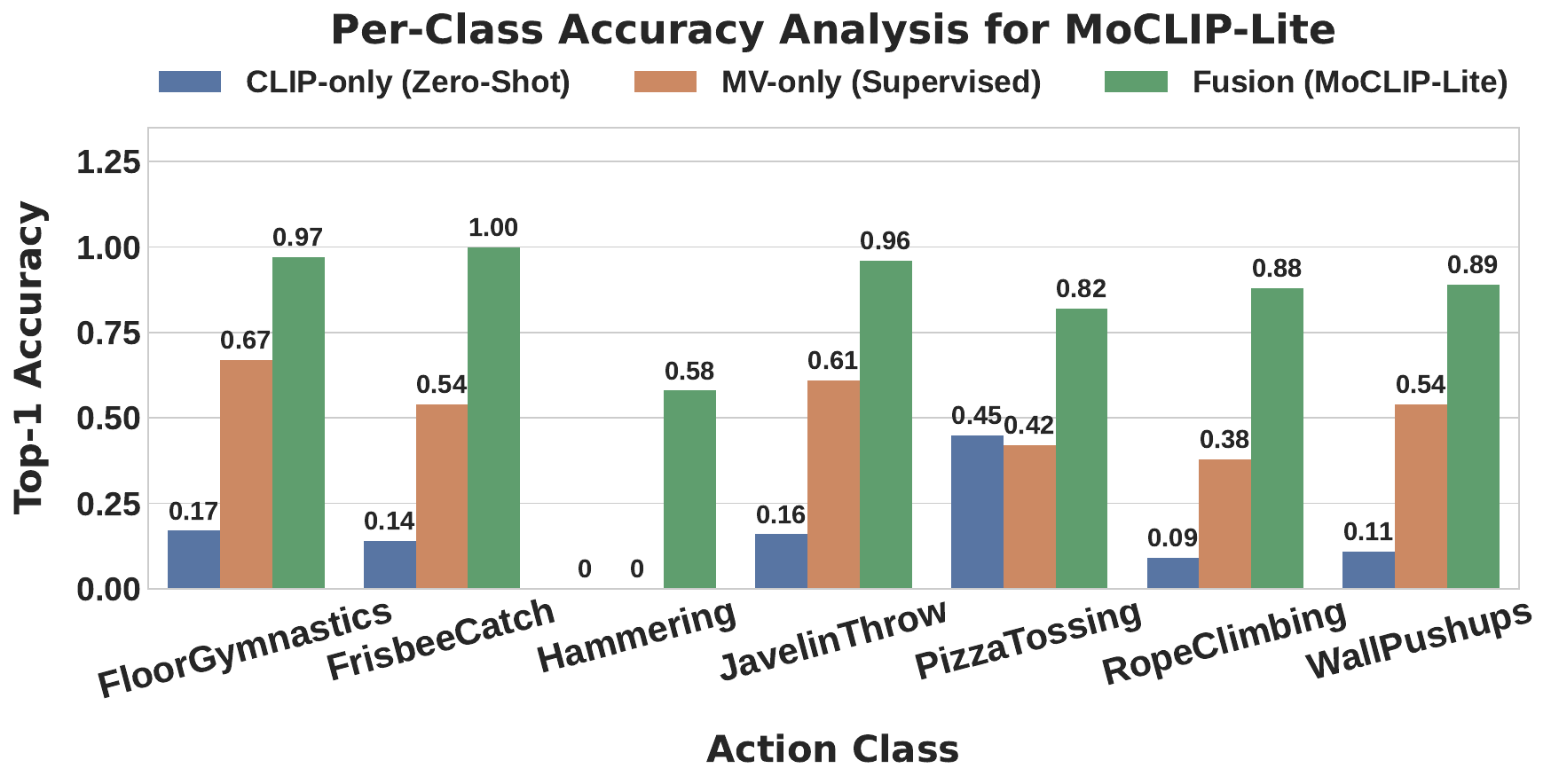}
    \caption{Per-class accuracy comparison on representative UCF101 classes. The MoCLIP-Lite consistently outperforms the best single-modality baseline, with particularly large gains on classes that are ambiguous to either modality alone.}
    \label{fig:class_analysis}
\end{figure}

\paragraph{Per-Class Analysis}
We analyzed the per-class accuracy to understand the complementary effects of fusion, with key examples visualized in Figure~\ref{fig:class_analysis}.  
For motion-dominated actions such as ``FloorGymnastics'', ``FrisbeeCatch'', ``JavelinThrow'', ``RopeClimbing'', and ``WallPushups'', the MV-only model already performs well, but MoCLIP-Lite further boosts performance from 0.44–0.67 up to 0.88–1.0. 
For complex actions requiring both appearance and motion understanding, such as "Hammering", both single-modality baselines fail (=0), whereas MoCLIP-Lite reaches 0.58. 
Finally, for visually driven actions like "PizzaTossing", the CLIP-only model performs slightly better than MV-only (0.45 vs. 0.42), yet MoCLIP-Lite still achieves a substantial improvement to 0.82, demonstrating the robustness and generality of our approach.

\subsection{Qualitative Results}
\begin{figure*}[ht!]
  \centering
  \subfloat[Archery]{\includegraphics[width=0.32\linewidth]{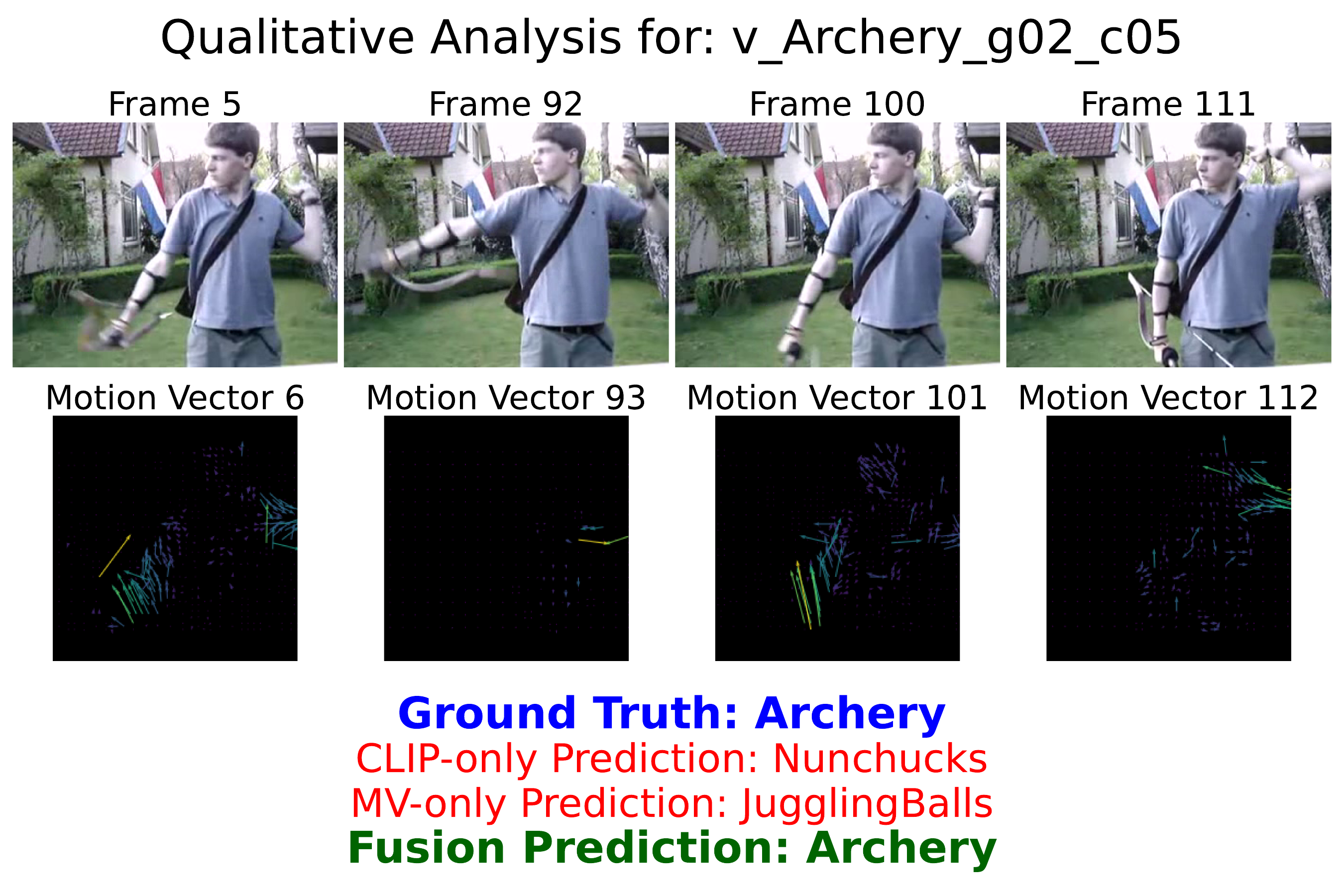}}\hfill
  \subfloat[BodyWeightSquats]{\includegraphics[width=0.32\linewidth]{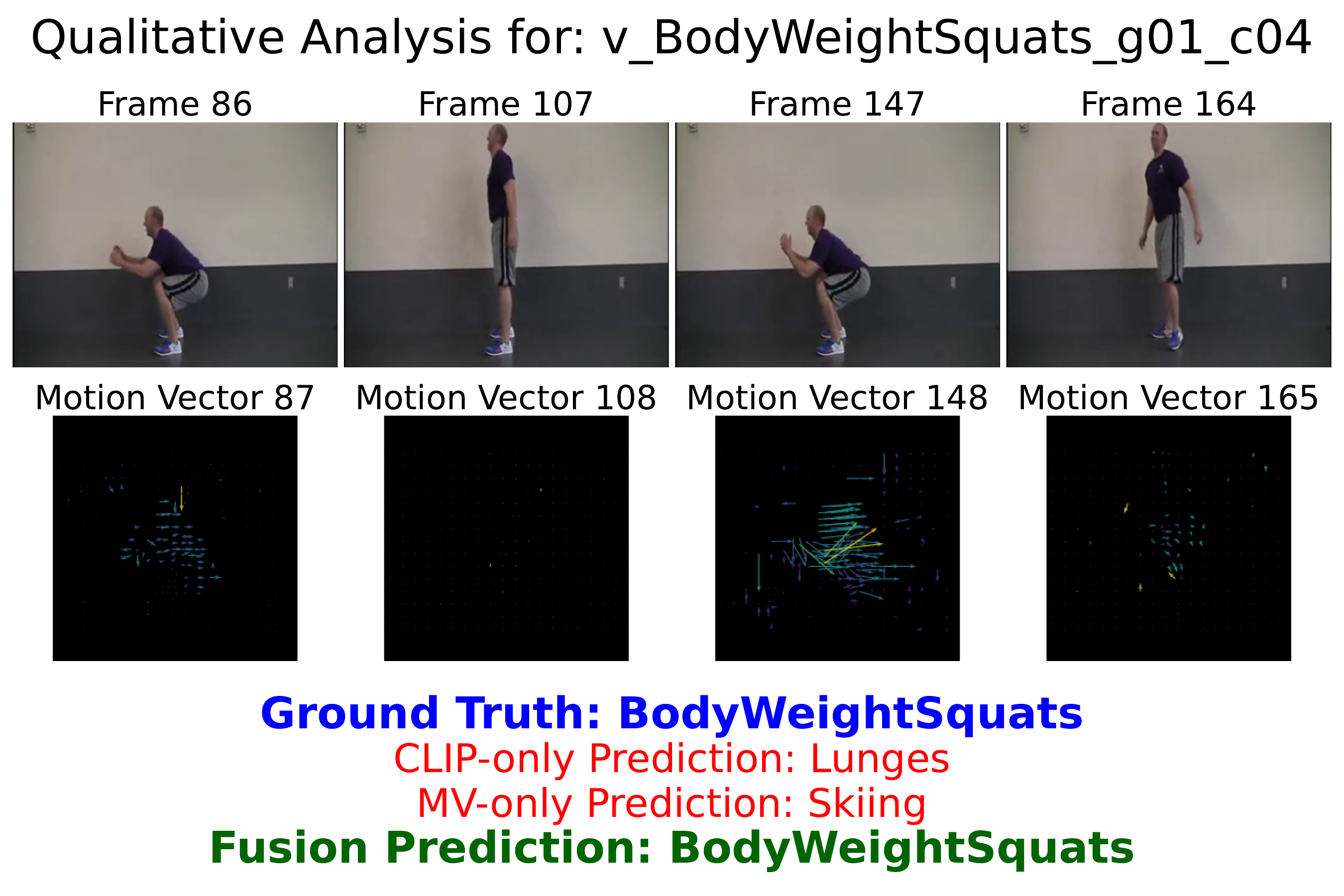}}\hfill
  \subfloat[BoxingPunchingBag]{\includegraphics[width=0.32\linewidth]{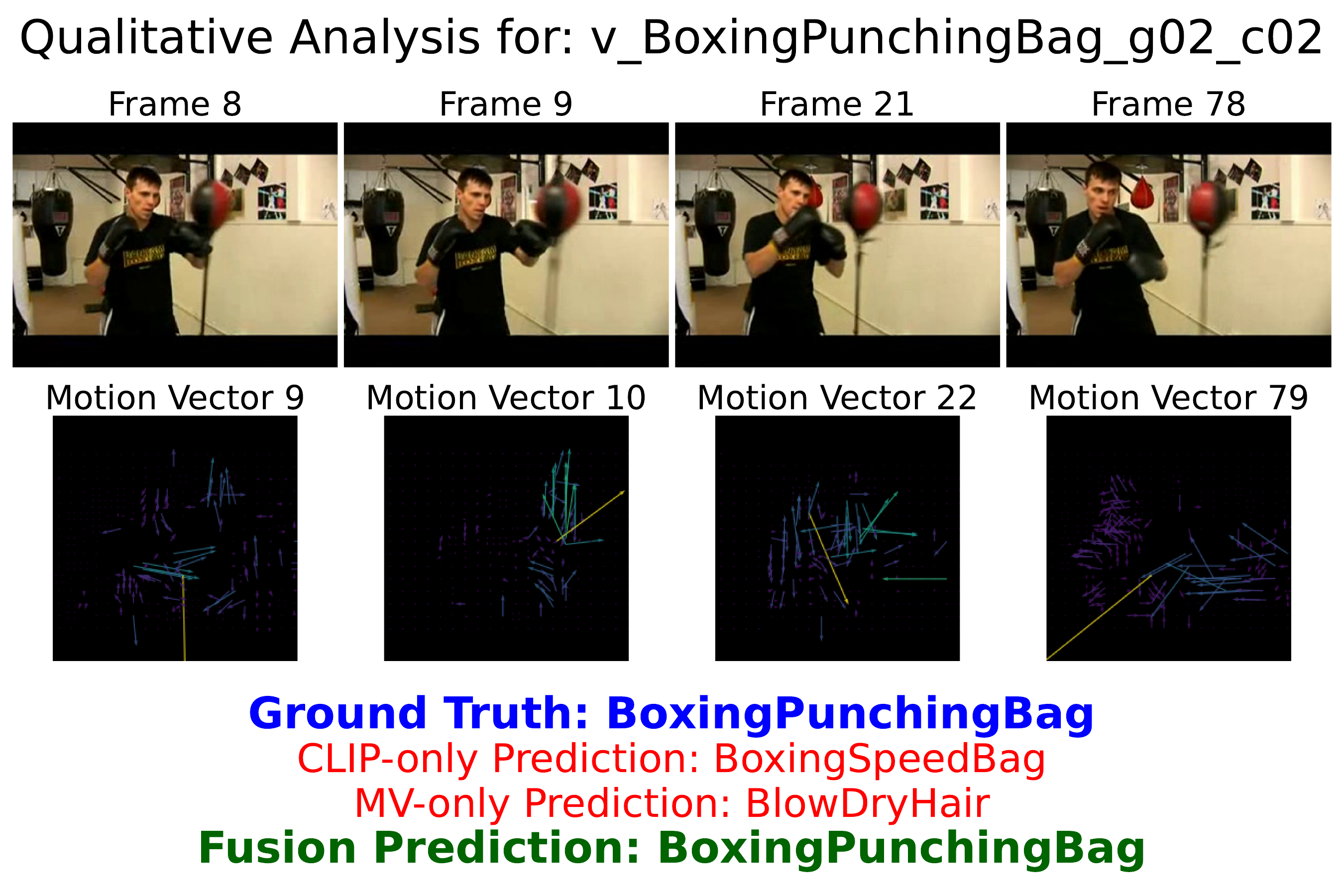}}\hfill
  \subfloat[CuttingInKitchen]{\includegraphics[width=0.32\linewidth]{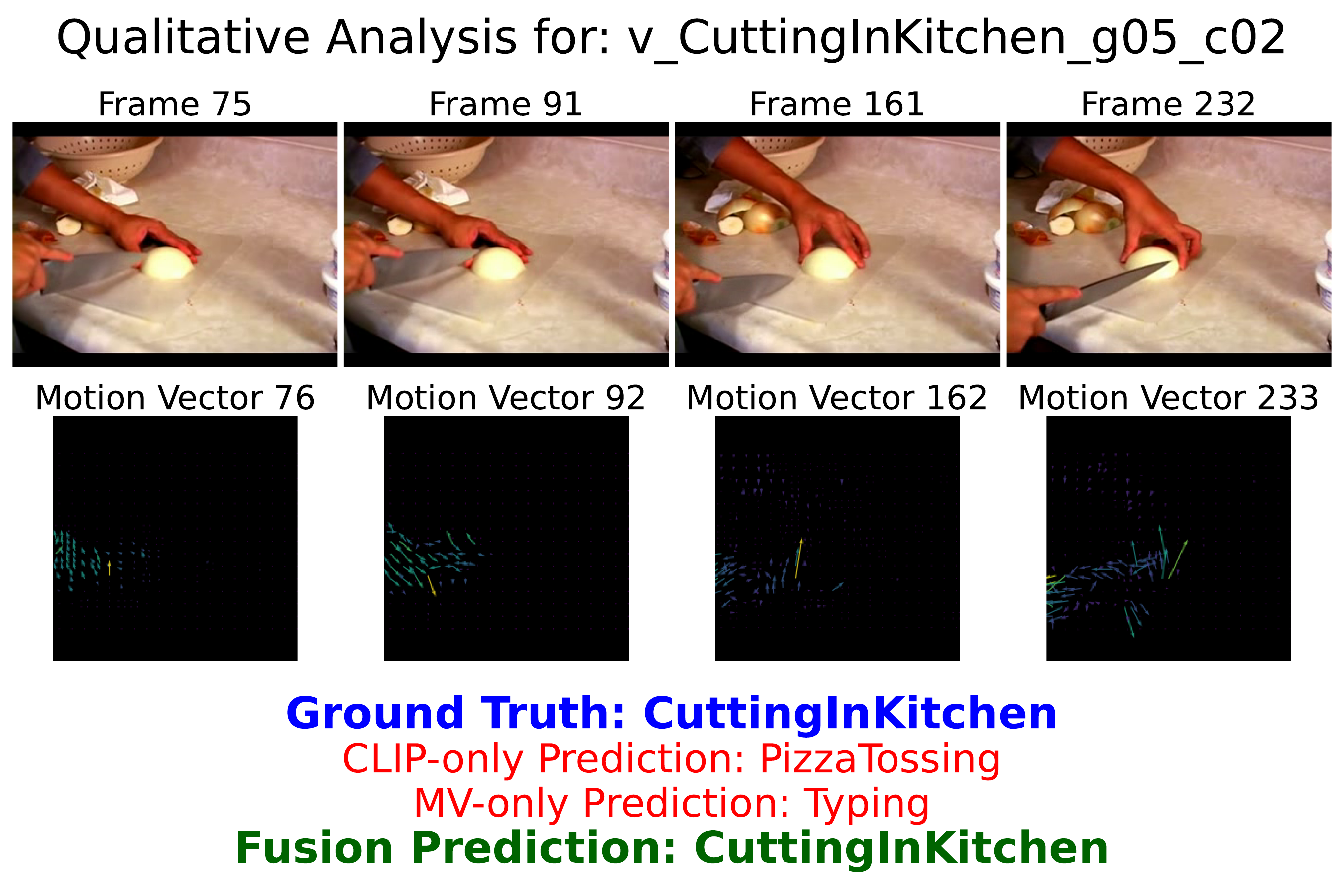}}\hfill
  \subfloat[FloorGymnastics]{\includegraphics[width=0.32\linewidth]{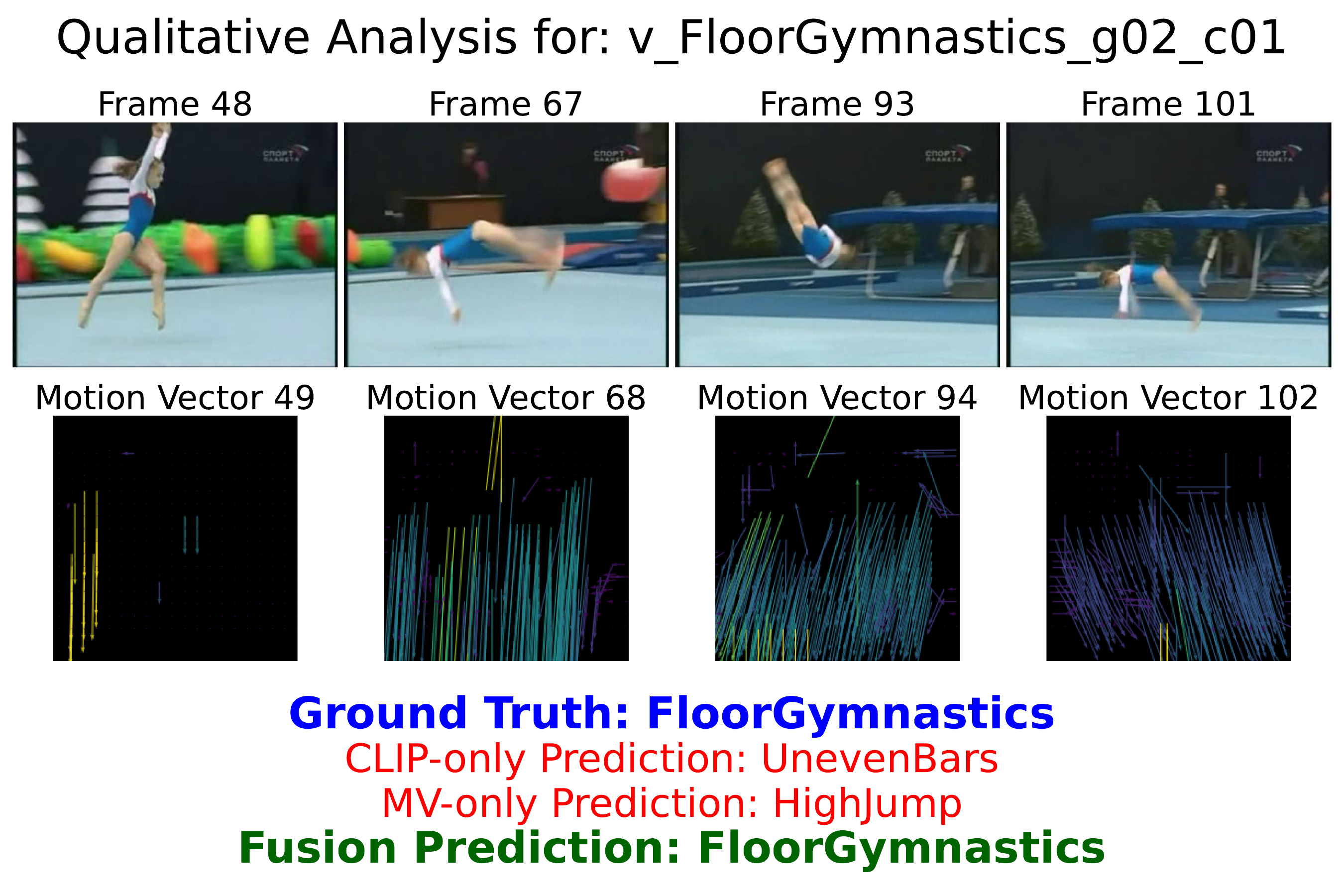}}\hfill
  \subfloat[FrontCrawl]{\includegraphics[width=0.32\linewidth]{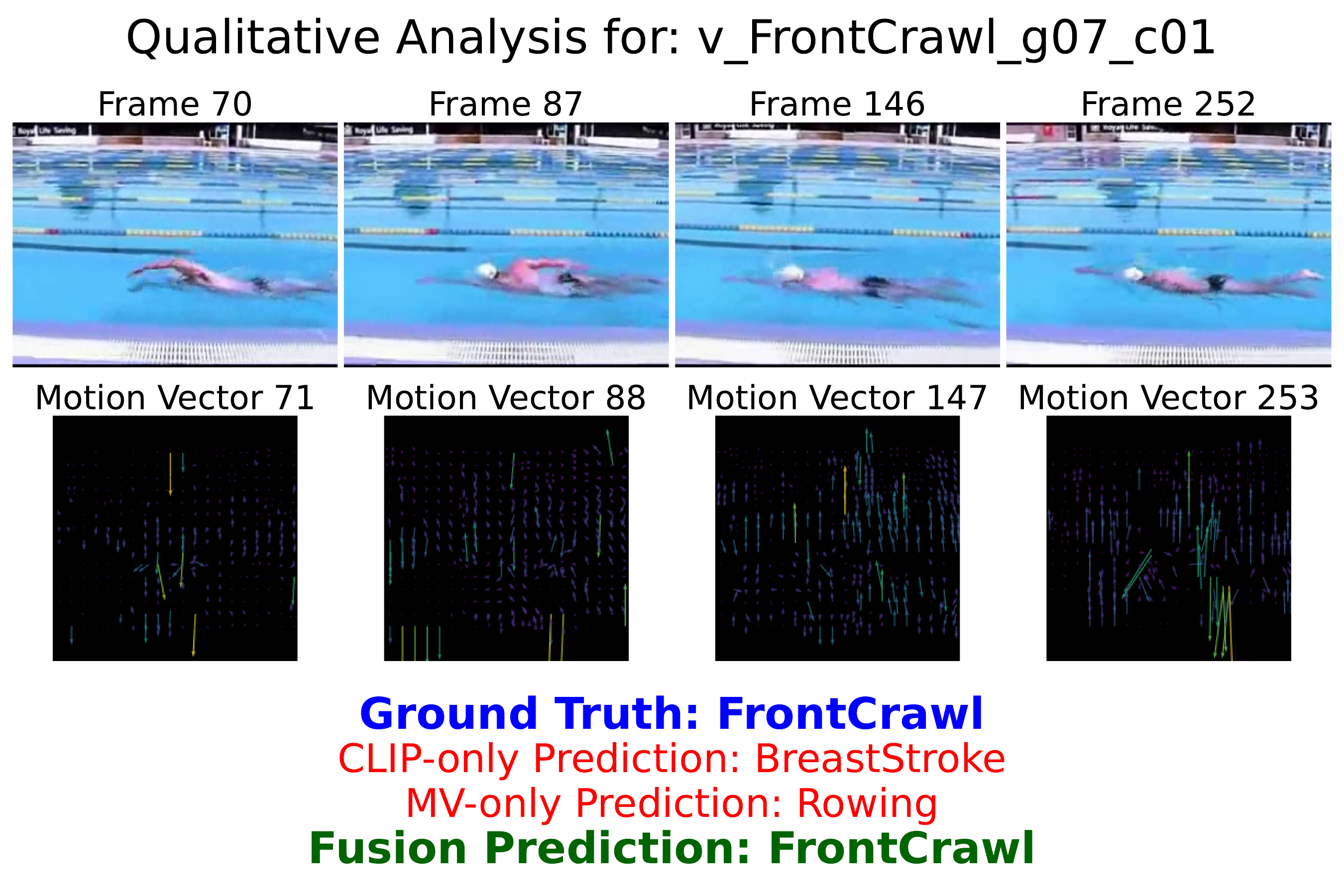}}
  
  \caption{Qualitative examples across different UCF101 classes, showing how fusing semantic and motion cues resolves ambiguity.}
  \label{fig:qual_examples_24}
\end{figure*}

To further illustrate the behavior of our model, we show representative qualitative examples in Figure~\ref{fig:qual_examples_24}. Taking the action “Archery” as an example, the CLIP-only model, influenced by visual context and object resemblance, incorrectly predicts “Nunchucks”. The MV-only model, which emphasizes localized motion cues, mistakes the repetitive arm movements for “Juggling Balls”. In contrast, MoCLIP-Lite integrates complementary evidence from both appearance and motion, successfully identifying the action as “Archery”. This example highlights the robustness of our framework in resolving ambiguities that arise when relying on a single modality.


%% file: sec/5_con.tex
\section{Conclusion and Future Work}
\label{sec:conclusion}

In this paper, we introduced MoCLIP-Lite, a highly efficient two-stream late fusion framework for video action recognition. We clearly showed that by combining the rich pre-computed semantic features from a frozen CLIP image encoder with dynamic motion cues from a lightweight supervised motion vector network, we can achieve strong and consistent performance. Our model reaches 89.2\% accuracy on UCF101, notably surpassing strong single-modality baselines while adding only negligible computational overhead during inference. This work establishes a practical and effective baseline, demonstrating that competitive results are possible without training large-scale video models from scratch.

Our method has some limitations. Relying on a single frame for the appearance pathway may miss actions where the key object is only briefly visible. Furthermore, our simple concatenation-based fusion, though efficient, may be weaker than attention-based mechanisms for integrating the two modalities. We leave these for future work.  

Future work includes three directions. First, incorporating features from multiple frames to provide a more robust and reliable semantic representation. Second, exploring more advanced fusion mechanisms, such as attention- or transformer-based layers, which could yield further gains. Finally, applying our efficient framework to larger and more diverse datasets would better validate its generalization ability.